%% file: main.tex
\documentclass[10pt,twocolumn,letterpaper]{article}

\usepackage{cvpr}            

\input{preamble}

\definecolor{cvprblue}{rgb}{0.21,0.49,0.74}
\usepackage[pagebackref,breaklinks,colorlinks,citecolor=cvprblue]{hyperref}

\def\paperID{9859} 
\def\confName{CVPR}
\def\confYear{2024}

\title{Cross-Modal Coordination Across a Diverse Set of Input Modalities}

\author{Jorge S\'anchez, \quad Rodrigo Laguna \\
MercadoLibre, Inc.\\
Argentina\\
{\tt\small \{jorge.sanchez, rodrigo.laguna\}@mercadolibre.com}
}

\begin{document}
\maketitle
\input{sec/0_abstract}    
\input{sec/1_intro}
\input{sec/2_losses}
\input{sec/3_experiments}
\input{sec/4_conclusions}
{
    \small
    \bibliographystyle{ieeenat_fullname}
    \bibliography{biblio}
}

\input{sec/5_supplementary}

\end{document}

%% file: preamble.tex
\usepackage{amsmath}
\usepackage{stmaryrd}
\usepackage{graphicx}
\usepackage{booktabs, multirow} 
\usepackage{caption}
\usepackage{subcaption}
\usepackage[table]{xcolor} 

\usepackage[normalem]{ulem}
\usepackage{xcolor}
\usepackage{soul}

\usepackage{microtype}

%% file: sec/0_abstract.tex
\begin{abstract}

Cross-modal retrieval is the task of retrieving samples of a given modality by using queries of a different one. Due to the wide range of practical applications, the problem has been mainly focused on the vision and language case, \eg text to image retrieval, where models like CLIP have proven effective in solving such tasks. The dominant approach to learning such coordinated representations consists of projecting them onto a common space where matching views stay close and those from non-matching pairs are pushed away from each other. Although this cross-modal coordination has been applied also to other pairwise combinations, extending it to an arbitrary number of diverse modalities is a problem that has not been fully explored in the literature. In this paper, we propose two different approaches to the problem. The first is based on an extension of the CLIP contrastive objective to an arbitrary number of input modalities, while the second departs from the contrastive formulation and tackles the coordination problem by regressing the cross-modal similarities towards a target that reflects two simple and intuitive constraints of the cross-modal retrieval task. We run experiments on two different datasets, over different combinations of input modalities and show that the approach is not only simple and effective but also allows for tackling the retrieval problem in novel ways. Besides capturing a more diverse set of pair-wise interactions, we show that we can use the learned representations to improve retrieval performance by combining the embeddings from two or more such modalities. 

\end{abstract}

%% file: sec/1_intro.tex
\section{Introduction}
\label{sec:intro}

Generating agents that can interact with the world, requires that they are able to perceive the environment in which they act. This environment is dynamic and populated with other agents with goals and constraints. The external world (to the agent) imposes constraints on the stimuli perceived by the agents, which helps to interrelate them in the context of the task the perceiving agent is willing to solve. Such constraints arise from the fact that the agent perceives the world concurrently in different ways, \eg using visual, auditory, and/or haptic information. These different sources of information are \emph{coordinated}, in the sense that they relate different perceptual stimuli to a common external event that is recognized as a single entity. This coordination between different and heterogeneous views of the same phenomenon can be regarded as one of the most important problems in building perceptual machines. From an application perspective, the problem of perceptual coordination is also crucial, as it would help the development of techniques to process the increasing amounts of multi-sensory digital information we are exposed to on a daily basis. This has driven an increasing interest in multimodal techniques \cite{liang2022fat,yin2023mmllm,xu2023multimodal}. However, most approaches studied and proposed in the literature reduce the multimodal learning problem only to two modalities. This choice is not arbitrary and can be seen as a consequence of the difficulties of generating reliable data for training such models since the same entity has to be sampled concurrently from the different views or modalities that define the problem (coordination constraint). Here, the use of vision, either in the form of still images or video, and language modalities has prevailed in the literature \cite{radford2021clip,jia2021align,li2022blip,sun2019videobert}. Other combinations, such as vision and audio \cite{morgado2021audio,shih2023speechclip,sanabria2021talk}, pose \cite{jiang2022text2human,huang2022imfnet}, attributes \cite{wang2021multimodal,yan2023learning}, among others, have also been explored. Nevertheless, the abundance of (weakly aligned) image and textual data paved the way for training high-capacity models at scale, proving such models to be effective in solving a variety of tasks. In our work, we aim to formalize a learning framework that allows us to 
learn coordinated representations across a possibly large and diverse set of modalities
, ranging from those that require complex encoders such as vision, language, and speech, to those captured by simple (learned or handcrafted) embeddings that are commonplace in many real-world applications. Equipped with such a framework we show we can apply the learned representation in novel ways, extending the capabilities of the bi-modal approaches commonly found in the literature. 

Our main contributions are the following:
\begin{itemize}
    \item We propose two different formulations for learning coordinated representations, the first based on an extension of the CLIP loss to an arbitrary set of pairwise combinations, and the second based on regressing the pairwise cross-similarities towards two intuitive constraints while accounting for the imbalance of matching and non-matching samples in the batch.
    \item We experimentally show that our approach competes favorably with specialized bi-modal approaches in two challenging datasets. More still, we are able to learn models that account for all pair-wise interactions in a simple manner.
    \item We show that by combining different modalities we can obtain large improvements in problems such as zero-shot classification and cross-modal retrieval.
\end{itemize}

The paper is organized as follows: in \cref{sec:prev_work} we discuss related work, in \cref{sec:losses} we introduce two different approaches for learning coordinated representations, in \cref{sec:experiments} we show experimental results under different settings for two challenging datasets. Finally, in \cref{sec:conclusions} we draw some conclusions.

\section{Related work\label{sec:prev_work}}

Multimodal learning is a topic that encompasses many different subjects within the machine learning literature. Here, we focus on methods that aim at learning generic multimodal representations. For a deeper and more comprehensive treatment of the topic, see \cite{yin2023mmllm,xu2023multimodal,akkus2023multimodal}.

Beyond the particular choice of input modalities, a first distinction of the different multimodal learning approaches in the literature relies on the way they combine such diverse inputs. Models like VisualBERT \cite{li2019visualbert} or LXMERT \cite{tan2019lxmert}, just to name a few, integrate these modalities via cross-modal fusion. While this approach is effective in solving a variety of vision and language tasks, it is difficult to scale to a larger set of input modalities, either because they would impose architectural constraints that are difficult to satisfy, or just because the interleaved nature of the fusion strategy narrows their application to problems that involve all modalities at once. In this regard, the CLIP \cite{radford2021clip} offers some advantages. On the one hand, and as we show in this paper, the model offers a simple way to fuse different input modalities. On the other, each input modality is encoded independently of the others (no cross-modal fusion) which enables the use of the different encoders either in isolation or combined. This, from an applications perspective, might be advantageous. Focusing on CLIP, another relevant factor is the possibility to choose different encoders for the input modalities. Being a training formalism, it does not interfere with the type of information the system is shown, as long as it remains consistent during the training. This has the practical advantage of not only being able to combine different families of models (transformer and CNN-based backbones, raw embeddings, etc.) but also offers a simple way to leverage powerful models pre-trained on a single modality, \eg the SpeechCLIP \cite{shih2023speechclip} model leverages three powerful transformers to coordinate image, speech, and textual information. This is also the rationale in models like \cite{girdhar2023imagebind}, where a strong pre-trained image encoder is used as an ``anchor" with the objective of mapping the different modalities into a common representation space.
Coordinated representations allow us not only to deal with problems that by their nature are intrinsically multi-modal (generation \cite{xie2023edit,galatolo2021generating}, VQA \cite{song2022clip,zakari2022vqa}, etc.) but also allow to expand the application domain for which these models were originally trained. One clear example is zero-shot classification \cite{xian2017gbu}, where aligning the image modality in a feature space that encodes semantic relations between the different categories (\eg text or attributes space), allows us to tackle such a problem as a cross-modal retrieval task. Our approach to learning coordinated representations overcomes these problems by providing an effective formulation. 

%% file: sec/2_losses.tex
\section{Similarity-based feature coordination}
\label{sec:losses}

This section describes our approach to learning aligned representations from an arbitrary set of input modalities. Let us assume we have $M$ different \emph{views} for some entity of interest, \eg visual scenes captured by different sensors (RGB, sonar, etc), products in an online catalog showing images, descriptive texts, and even audio transcriptions, etc. For simplicity, let us also assume that each modality is independently encoded into a vector $x^{(m)}\in\mathcal{M}_m$ of dimensionality $D^{(m)}$, $m=1,\dots, M$. Our goal is to learn, for each of such representations, a mapping into a common $D$-dimensional space so that different views from the same entity lead to similar vector embeddings under a suitable metric. Let $f_{\theta_m}:\mathbb{R}^{D^{(m)}}\rightarrow\mathbb{R}^D$ denote the mapping corresponding to the $m$-th such modality, with parameter vector $\theta_m$. 
Given a training set with $N$ samples and modalities $m_i$, $i=1, \dots, M$, learning is performed by minimizing a suitable loss defined over mini-batches of size $B$, as:
\begin{equation}\label{eq:loss_generic}
    \mathcal{L}(\theta_1,\dots,\theta_M) = \sum_{k=1}^{\lfloor N/B\rfloor} \sum_{\substack{i,j\\i<j}} \ell(f_{k_1:k_2}^{(m_i)}, f_{k_1:k_2}^{(m_j)}; \theta_i,\theta_j)
\end{equation}
where $f_{k_1:k_2}^{(m_i)}$ denotes the samples from modality $m_i$ in the $k$-th mini-batch, \ie samples with indices ranging from $k_1=(k-1)B+1$ to $k_2=kB$ inclusive. 

Let us now define $S^{(m_i, m_j)}$ as the matrix of pairwise similarities between the representations of modalities $m_i$ and $m_j$ for the samples in the mini-batch. This is a $B \times B$ real-valued matrix whose $pq$ element encodes the similarity between the projections of the representations for modalities $m_i$ and $m_j$ of samples $p$ and $q$, respectively, as:
\begin{equation}\label{eq:s_matrix}
    S^{(m_i, m_j)}_{pq} = sim\left(f_p^{(m_i)}, f_q^{(m_j)}\right),
\end{equation}
where $sim:\mathbb{R}^D\times \mathbb{R}^D\rightarrow\mathbb{R}$ measures the compatibility between views $f_p^{(m_i)}$ and $f_q^{(m_j)}$. If we choose $sim(\cdot, \cdot)$ to be the cosine similarity, $S^{(m_i, m_j)} \in {\left[-1,1\right]}^{B\times B}$.
Also, we have $S^{(m_i, m_j)}={(S^{(m_j,m_i)})}^T$. We can redefine \cref{eq:loss_generic} as:
\begin{equation}\label{eq:loss}
    \sum_{k=1}^{\lfloor N/B\rfloor} \sum_{\substack{i,j\\i<j}} \ell(S^{(m_i, m_j)}_{k_1:k2}; \theta_i, \theta_j)\ .
\end{equation}
This formulation allows us to generalize CLIP \cite{radford2021clip} to multiple modalities. If we set $M=2$ (\emph{v}: vision, \emph{l}: language) and $\ell$ to the symmetric cross-entropy loss:
\begin{equation}\label{eq:clip_loss}
    \ell(S^{vl};\theta_v,\theta_l) = \frac{1}{2}\left(\ell_{\text{CE}}(S^{vl};\theta_v,\theta_l) + \ell_{\text{CE}}({S^{lv}};\theta_v,\theta_l)\right)\ ,
\end{equation}
we recover the original CLIP formulation, where in this case, $\ell_{\text{CE}}(S)$ is defined as:
\begin{equation}\label{eq:ce_loss}
    \ell_{\text{CE}}(S) = - \sum_p \log \frac{\exp{(\tau S_{pp})}}{\sum_{q}\exp{(\tau S_{pq})}}\ ,
\end{equation}
and $\tau$ is a (fixed or learned) temperature parameter. 

For $M>2$, plugging this loss into \cref{eq:loss} leads to a summary loss that takes into account the $\binom{M}{2}$ all possible pairwise combinations between $M$ modalities. The goal of this loss is thus to maximize the similarity between the different views of each training instance while minimizing the similarities to other views of the non-matching samples in the batch. 

Note that, although the combination of the loss terms in \cref{eq:loss} is linear, they are not independent since minimizing the loss for a given pair must account also for the interaction of these modalities with the $(M-2)$ remaining ones. 

In what follows, we refer to the multimodal extension of the CLIP loss as \emph{pairwise cross-modal contrastive} (PCMC).


\subsection{Non-contrastive coordination}

In this section, we provide an alternative formulation to the contrastive approach outlined before. Inspired by \cite{zbontar2021barlow}, we look at the different modalities as jointly distributed random variables and seek to maximize (minimize) the correlation between matching (non-matching) sample pairs. Instead of computing the Pearson coefficient explicitly as in \cite{zbontar2021barlow}, we look at the constraints that should be satisfied by the pairwise scores \cref{eq:s_matrix}. If we assume that the embeddings for each modality are normalized for zero-mean\footnote{This is achievable if we add a LayerNorm layer at the end of the projection head of every encoder $f^{(p)}$.} ($\bar{x}^{(p)}_i\approx 0$ for every $p$), the Pearson correlation equals the normalized dot-product (cosine score). In this case, we can formulate two simple and intuitive constraints:
\begin{enumerate}
    \item matching pairs should have a score close to one,
    \item non-matching pairs should be uncorrelated, \ie they should have a score close to zero.
\end{enumerate}
We can enforce these constraints with the following loss:
\begin{equation}\label{eq:reg_loss}
    \ell_{\text{R}}(S^{(m_i,m_j)}) = {\|S^{(m_i,m_j)} - T\|}_F^{2+\rho} \quad ,
\end{equation}
where ${\|\cdot\|}_F$ denotes the Frobenius-norm and $T$ is a target matrix. Although a canonical choice for this matrix is the identity, \ie $T=\mathrm{I}$, some problems require some additional considerations. For instance, there might be the case that two samples share some of the views, \eg this is common in captioning datasets where different image-text pairs share the same image. In our case, we set the entries of this matrix as follows:
\begin{equation}
    T_{pq} = \llbracket \max_{i=1,\dots,M}{S_{pq}^{(m_i,m_i)}} > t \rrbracket \quad,
\end{equation}
\ie by looking at the maximum similarity between the same view of each sample pair in the batch. We set the threshold $t$ to a high value relative to the metric to ensure the match is correct. We use $t=0.99$ for the cosine score.

The power-modulating factor $\rho$ in \cref{eq:reg_loss} is included to balance the proportion of matching to non-matching samples \cite{lu2018shrinkage}. In practice, we use $\rho=1$. 

In what follows, we refer to this formulation as \emph{pairwise cross-modal regression} (PCMR).

\subsection{Departing from the fully-aligned case}

In the case where not all samples share the same set of input modalities, as in \cite{girdhar2023imagebind}, we can modify \cref{eq:clip_loss,eq:reg_loss} to include a mask that prevents unpaired samples from contributing to the loss. For the PCMC approach, we can adaptively mask the similarities element-wise, as follows:
\begin{equation}\label{eq:mask_c}
    {S'}^{(m_i,m_i)} = H^C + S^{(m_i,m_i)}
\end{equation}
with $H^C_{pq} = 0$ if $m_i$ and $m_j$ are present in samples $p$ and $q$, respectively, and $-\infty$ otherwise. This masking operation acts in conjunction with the softmax operation in \cref{eq:ce_loss}, avoiding the penalization of missing cross-modal pairs.

For the PCMR approach, we can change \cref{eq:reg_loss} to:
\begin{equation}\label{eq:mask_c}
    \ell'_{\text{R}}(S^{(m_i,m_j)}) = {\|H^R \odot (S^{(m_i,m_j)} - T)\|}_F^{2+\rho} \quad ,
\end{equation}
with $H^R_{pq} = 1$ if $m_i$ and $m_j$ are present in samples $p$ and $q$, respectively, and $0$ otherwise.  $\odot$ denotes element-wise product. 

Finally, note that different from \cite{girdhar2023imagebind}, we do not require the image or any other modality to act as an ``anchor", \ie a modality that has to be present in all pairwise alignment sub-problems.

%% file: sec/3_experiments.tex
\section{Experiments}
\label{sec:experiments}



\paragraph{Datasets.} We run experiments in two different datasets with different modality combinations: Flickr8k Audio Captions Corpus \cite{harwath2015f8k}, and CUB \cite{wah2011cub}. 

Flickr8k consists of 8k images paired with 5 different textual and spoken captions each. There is a total of 46 hours of speech. We evaluate cross-modal retrieval performance and report recall@1 (r@1) and recall@5 (r@5) metrics. This dataset accounts for 3 different modalities (I: Image, T: text, S: speech), resulting in 3 different pairwise combinations.

For CUB, we use the CUB-Captions variant proposed in \cite{chun2021probabilistic}, which consists of 11788 images from 200 fine-grained bird classes, together with 10 different captions per image by \cite{reed2016learning}. Besides the image and text modalities, we also consider attribute and class embeddings as additional modalities. Class embeddings are built by averaging the instance-level attribute vectors from each class. We use the 312-dimensional ``continuous" attributes provided with the dataset. We follow \cite{chun2021probabilistic} and use 150 classes for training and validation, and leave 50 for testing. Besides providing a more challenging setup, this also allows us to test zero-shot classification performance as a cross-modal task. For this dataset, in addition to the recall@1 metric, we also report the R-Precision score (R-P) proposed by \cite{musgrave2020rp} which measures, for each query, the proportion of positives in the top-$r$ retrieved items, with $r$ the number of true matches. This dataset accounts for 4 different modalities (I: image, T: text, A: attributes, C: class), resulting in 6 pairwise combinations.

Besides the number of input modalities, an important difference between these datasets relies on the different "granularities" at which they encode different aspects of the input. For instance, we have 10 different spoken and written captions for each image in the Flickr8k dataset. Different captions describing the same image can be thought of as finer-resolution representations compared to the image they describe. For CUB, this is more pronounced, as we have 5 different captions per image, a coarse attribute descriptor for each such image, and aggregated class descriptors that encode higher-level class-level abstractions. As we will see in the experiments, such diversity makes learning coordinated representations especially challenging.


\paragraph{Model design and training strategy.} For all modalities, we follow the same encoding strategy which consists of using a linear layer to project the input embeddings (either raw fetures or computed by the backbone network) onto a common space of $D=256$ dimensions. We use these features to feed a small feed-forward subnetwork consisting of a single hidden dimension (dimensionality of $256$), a residual connection, and a LayerNorm layer at the end, similar to the one used in the encoder block of the transformer architecture.
To train our models, we use a learning rate of $10^{-4}$ for the projections and $10^{-6}$ for the pre-trained backbones (if not frozen), and a weight decay value of $0.2$. We train our models for a maximum of $50$ epochs using a cosine schedule and the Adam optimizer. We use a batch size of $80$ for Flickr8k and $128$ for CUB. We monitor the average cross-modal performance on the validation set and stop training if there is no improvement after $5$ consecutive epochs. 
We do not apply any particular dataset-specific fine-tuning. For the image and text modalities, we use the ViT/B-32 \cite{dosovitskiy2020vit} and BERT-like \cite{kenton2019bert} encoders from CLIP \cite{radford2021clip}. For speech, we use HuBERT-Base \cite{hsu2021hubert} with a weighted pooling of the model's hidden states as described in \cite{yang2021superb}. For the attribute and class embeddings, we use the precomputed features provided with the datasets as described before. 
All experiments were run using a single V100 GPU with 16G of RAM. In all cases, we use a standard image augmentation strategy (as implemented in the \textrm{timm} \cite{rw2019timm} library), and a text augmentation strategy based on EDA \cite{wei2019eda} as described in \cite{fini2023improved}.

\subsection{Cross-modal retrieval and model design}
\label{sec:cross_modal_results}
\begin{table*}
\small
\centering
\resizebox{\linewidth}{!}{

\begin{tabular}{llccccccccccccc}\toprule
\multicolumn{2}{l}{\multirow{2}{*}{\textbf{Model}}} &\multicolumn{6}{c}{\textbf{r@1}} &\multicolumn{6}{c}{\textbf{r@5}} \\
\cmidrule(l{0pt}r{3pt}){3-8} \cmidrule(l{3pt}r{0pt}){9-14}
& &I$\rightarrow$T &T$\rightarrow$I &I$\rightarrow$S &S$\rightarrow$I &T$\rightarrow$S &S$\rightarrow$T &I$\rightarrow$T &T$\rightarrow$I &I$\rightarrow$S &S$\rightarrow$I &T$\rightarrow$S &S$\rightarrow$T \\\midrule
\multirow[t]{3}{*}{MILAN \cite{sanabria2021talk}} &I+S &- &- &49.6 &33.2 &- &- &- &- &79.2 &62.7 &- &- \\
&I+S$\rightarrow$ASR$\rightarrow$T &63.0 &46.9 &- &- &- &- &- &- &- &- &- &- \\
&I+T &65.7 &52.1 &- &- &- &- &- &- &- &- &- &- \\
\multirow[t]{2}{*}{SpeechCLIP \cite{shih2023speechclip}} &Parallel &- &- &41.3 &26.7 &- &- &- &- &73.9 &57.1 &- &- \\
&Parallel large &- &- &54.5 &39.1 &22.5 &19.6 &- &- &84.5 &72.0 &44.1 &44.1 \\
\midrule
\multicolumn{2}{l}{Contrastive} &67.9 &54.9 &42.3 &32.4 &70.1 &78.1 &89.9 &84.3 &76.6 &64.6 &91.2 &94.2 \\
\multicolumn{2}{l}{Non-contrastive} &66.8 &55.8 &44.5 &34.8 &84.0 &88.2 &89.5 &84.1 &77.8 &66.8 &96.0 &98.0 \\
\bottomrule
\end{tabular}
}
\caption{Cross-modal retrieval performance on the Flickr8k dataset.}
\label{tbl:f8k}
\vspace{-1mm}
\end{table*}
\cref{tbl:f8k} shows cross-modal retrieval performance on the Flickr8 dataset, for the r@1 and r@5 metrics, where the notation X$\rightarrow$Y denotes using queries from modality ``X" to retrieve those from modality ``Y". We compare our PCMC and PCMR variants described in ~\cref{sec:losses} using frozen backbones for all three modalities since fine-tuning all three backbones (ViT/B-32, BERT, and HuBERT) is too memory expensive. Using frozen backbones, we are able to use a batch size $B=80$. From the table, we see that the PCMR formulation leads to better performance than PCMC overall, the only exceptions being the $I\rightarrow T$ subtask under the r@1 metric. We also tried cross-validate the $\rho$ parameter in \cref{eq:reg_loss}. By setting it to zero, we observed a decrease of $17\%$ on average, showing the importance of balancing the number of matches and non-matches similarities when learning the models. We did not observe any significant gain by setting this parameter to a different value, and we use $\rho=1$ in the rest of the paper.

The table also compares performance with two other models from the literature: MILAN \cite{sanabria2021talk} and SpeechCLIP \cite{shih2023speechclip}. MILAN is a dual encoder based on CPC-8k features \cite{kawakami2020cpc} and an EfficientNet-B4 image backbone, pre-trained on a large set of synthesized spoken captions using a masked softmax loss \cite{ilharco2019large}. We consider the following settings: MILAN trained on image and speech data (I+S), the same model but using an automatic speech recognition (ASR) system to transcribe the speech signal to written text (I+S$\rightarrow$ASR$\rightarrow$T), and a similar system using a BERT text encoder (I+T). As seen from the table, our models show competitive or better performance in all cross-modal tasks, except for I$\rightarrow$S, for which we observe a gap (+11.5\% and +1.8\% relative to PCMR for the r@1 and r@5 metrics, respectively). Note, however, that our models are able to capture all pairwise interactions. 
SpeechCLIP is based on frozen HuBERT and CLIP encoders where the speech projection head is trained contrastively. The ``Parallel" version of the model is based on a HuBERT-Base and ViT-B/32 backbones, while the ``Parallel Large" variant uses a HuBERT-Large and a ViT-L/14 encoder. We also report a supervised variant of the parallel large model, where the image backbone is replaced with the (pre-trained) text encoder from CLIP. The $T\rightarrow S$ and $S\rightarrow T$ of the Parallel Large model correspond to using the learned speech encoder together with the pre-trained text encoder from CLIP. From the results in \cref{tbl:f8k}, we see that both PCMC and PCMR perform better than the Parallel variant which employs the same backbone models, while they lag behind the Parallel Large variant that relies on more capable image and text encoders. We believe using larger backbones would provide a simple way to improve performance in our case, but not being the focus of the paper, we leave it to future work. Interestingly, we perform better than this model in text-to-speech and speech-to-text. Finally, besides the good performance observed by our models, we are able to tackle all cross-modal tasks simultaneously and consistently.

\begin{table*}
\small
\centering
\begin{tabular}{lccccccccccccc}\toprule
\multirow{2}{*}{\textbf{Model}} &\multicolumn{6}{c}{\textbf{R-P}} &\multicolumn{6}{c}{\textbf{r@1}} \\
\cmidrule(l{0pt}r{3pt}){2-7} \cmidrule(l{3pt}r{0pt}){8-13}
&I+T &I+C &I+A &T+C &T+A &C+A &I+T &I+C &I+A &T+C &T+A &C+A \\\midrule
PCME \cite{chun2021probabilistic} &26.6 &- &- &- &- &- &41.0 &- &- &- &- &- \\
DAA \cite{li2022differentiable} &28.4 &- &- &- &- &- &45.5 &- &- &- &- &- \\
PCMDA \cite{wang2022paired} &29.9 &- &- &- &- &- &46.7 &- &- &- &- &- \\
\midrule
PCMC & & & & & & & & & & & & \\
\hspace{10pt} frozen &\cellcolor[HTML]{fffffe}24.6 &61.6 &36.4 &26.9 &\cellcolor[HTML]{fffefe}19.2 &53.2 &39.0 &70.2 &53.4 &\cellcolor[HTML]{fbe9e7}45.1 &\cellcolor[HTML]{fef8f7}31.7 &72.2 \\
\hspace{10pt} frozen image$^\dag$ &\cellcolor[HTML]{e98980}29.3 &\cellcolor[HTML]{f8d6d3}62.9 &\cellcolor[HTML]{f7d0cc}36.9 &\cellcolor[HTML]{e77f77}35.3 &\cellcolor[HTML]{e67c73}24.0 &\cellcolor[HTML]{f5c5c0}55.1 &\cellcolor[HTML]{ec9992}45.9 &\cellcolor[HTML]{f6cac6}72.9 &\cellcolor[HTML]{f8d8d5}54.8 &\cellcolor[HTML]{ed9c95}49.2 &\cellcolor[HTML]{e67c73}38.7 &\cellcolor[HTML]{f3bbb6}74.0 \\
\hspace{10pt} frozen text &24.5 &\cellcolor[HTML]{f0aca6}64.4 &\cellcolor[HTML]{f2b5b0}37.3 &26.9 &19.2 &\cellcolor[HTML]{f1b3ae}55.5 &\cellcolor[HTML]{fae0de}41.2 &\cellcolor[HTML]{f1b2ac}74.1 &\cellcolor[HTML]{eea59e}56.7 &43.9 &31.3 &\cellcolor[HTML]{f4c3be}73.8 \\
\hspace{10pt} fine-tuned &\cellcolor[HTML]{e67c73}29.8 &\cellcolor[HTML]{e67c73}66.4 &\cellcolor[HTML]{e67c73}38.6 &\cellcolor[HTML]{e67c73}35.5 &\cellcolor[HTML]{e78077}23.9 &\cellcolor[HTML]{e67c73}56.4 &\cellcolor[HTML]{e67c73}47.6 &\cellcolor[HTML]{e67c73}76.5 &\cellcolor[HTML]{e67c73}58.1 &\cellcolor[HTML]{e67c73}51.2 &\cellcolor[HTML]{e77f76}38.5 &\cellcolor[HTML]{e67c73}75.8 \\
PCMR & & & & & & & & & & & & \\
\hspace{10pt} frozen &24.4 &\cellcolor[HTML]{9abbf2}49.5 &\cellcolor[HTML]{9abcf3}36.1 &24.4 &\cellcolor[HTML]{fcfdff}19.9 &\cellcolor[HTML]{f4f7fe}49.8 &38.2 &\cellcolor[HTML]{bfd4f7}60.0 &52.9 &\cellcolor[HTML]{f4f8fe}36.7 &\cellcolor[HTML]{fdfeff}32.2 &\cellcolor[HTML]{6d9eeb}71.2 \\
\hspace{10pt} frozen image &\cellcolor[HTML]{79a6ed}29.5 &\cellcolor[HTML]{6d9eeb}50.3 &\cellcolor[HTML]{6d9eeb}36.4 &\cellcolor[HTML]{6d9eeb}31.0 &\cellcolor[HTML]{6d9eeb}24.7 &\cellcolor[HTML]{a9c5f4}50.5 &\cellcolor[HTML]{93b7f1}45.3 &\cellcolor[HTML]{6d9eeb}62.3 &\cellcolor[HTML]{f8fbff}53.0 &\cellcolor[HTML]{6d9eeb}45.4 &\cellcolor[HTML]{6d9eeb}40.0 &\cellcolor[HTML]{d1e0fa}69.5 \\
\hspace{10pt} frozen text &24.4 &\cellcolor[HTML]{e2ecfc}47.8 &35.9 &\cellcolor[HTML]{fdfeff}24.5 &19.8 &\cellcolor[HTML]{6d9eeb}51.1 &\cellcolor[HTML]{e6eefc}40.4 &\cellcolor[HTML]{d1e0fa}59.4 &\cellcolor[HTML]{9cbdf2}55.3 &35.8 &32.0 &\cellcolor[HTML]{bcd2f7}69.9 \\
\hspace{10pt} fine-tuned &\cellcolor[HTML]{6d9eeb}29.9 &46.8 &\cellcolor[HTML]{eff5fe}35.9 &\cellcolor[HTML]{72a2ec}30.9 &\cellcolor[HTML]{73a2ec}24.5 &49.8 &\cellcolor[HTML]{6d9eeb}47.0 &57.8 &\cellcolor[HTML]{6d9eeb}56.5 &\cellcolor[HTML]{8fb5f0}43.4 &\cellcolor[HTML]{6e9fec}40.0 &68.2 \\
\bottomrule
\end{tabular}
\caption{Cross-modal retrieval performance on the CUB dataset. We use different shades of red and blue interpolated linearly between the min and max of each column and group to highlight performance ranks. $^\dag$ PCMC with a frozen image backbone resembles ImageBind \cite{girdhar2023imagebind}, where the image modality is used as an anchor for learning pairwise interactions.}
\label{tbl:cub}
\vspace{-1mm}
\end{table*}
\cref{tbl:cub} shows cross-modal retrieval results on the CUB dataset, in a four-modal setup. In this case, we report the average of the pairwise metrics due to space constraints. Full results can be found as supplemental material. We show performance for different configurations of frozen/fine-tuned image and language backbones and compare against recent models from the literature on cross-modal retrieval specialized for the image and text modalities. We were unable to find models for cross-modal retrieval that go beyond the image and language modality on this dataset. In our case, these are the only modalities that have specialized backbones since, for the class and attribute representations, we rely on pre-computed embeddings. For both formulations, we evaluate the effect of freezing/fine-tuning either or both the image and text backbone.

Overall, we observe better cross-modal performance when all backbones are being fine-tuned
. We note that freezing the text modality is the most detrimental alternative, showing the importance of the language modality in cross-modal tasks. However, freezing the image backbone does not seem too detrimental. This observation goes in line with the recent ImageBind model \cite{girdhar2023imagebind} that uses the image modality as an anchor for learning pairwise alignments independently with each other modality. Unlike Flickr8k, in this case, we observe consistently better performance for the contrastive over the non-contrastive formulation, perhaps due to the class-level nature of the problem and metrics involved, where contrastively pushing embeddings to be close or away from each other might bring some advantages from a class-level perspective.


Compared to the state-of-the-art, PCMC compares favorably with PCME \cite{chun2021probabilistic}, DAA \cite{li2022differentiable}, and PCMDA \cite{wang2022paired}. PCME uses a probabilistic formulation to learn parametric distributions in the embedding space. DAA introduces a differentiable objective with the goal of training robust models in noisy datasets. PCMDA uses a data augmentation approach based on the StyleGAN2 generative model. Again, our models not only compare favorably with these other strategies but allow us to capture a more diverse and interesting set of cross-modal interactions in a simple yet effective manner. Note that the data generation approach of PCMDA could also be used to improve the performance of our models. Since our goal is not to achieve the best possible performance but to show a reliable way to learn from multiple modalities, we leave this to be explored in future works.
\subsection{Does M-modal learning help pairwise retrieval?}
\begin{figure}
\centering
\begin{subfigure}[b]{0.5\textwidth}
    \centering
    \includegraphics[width=0.9\textwidth]{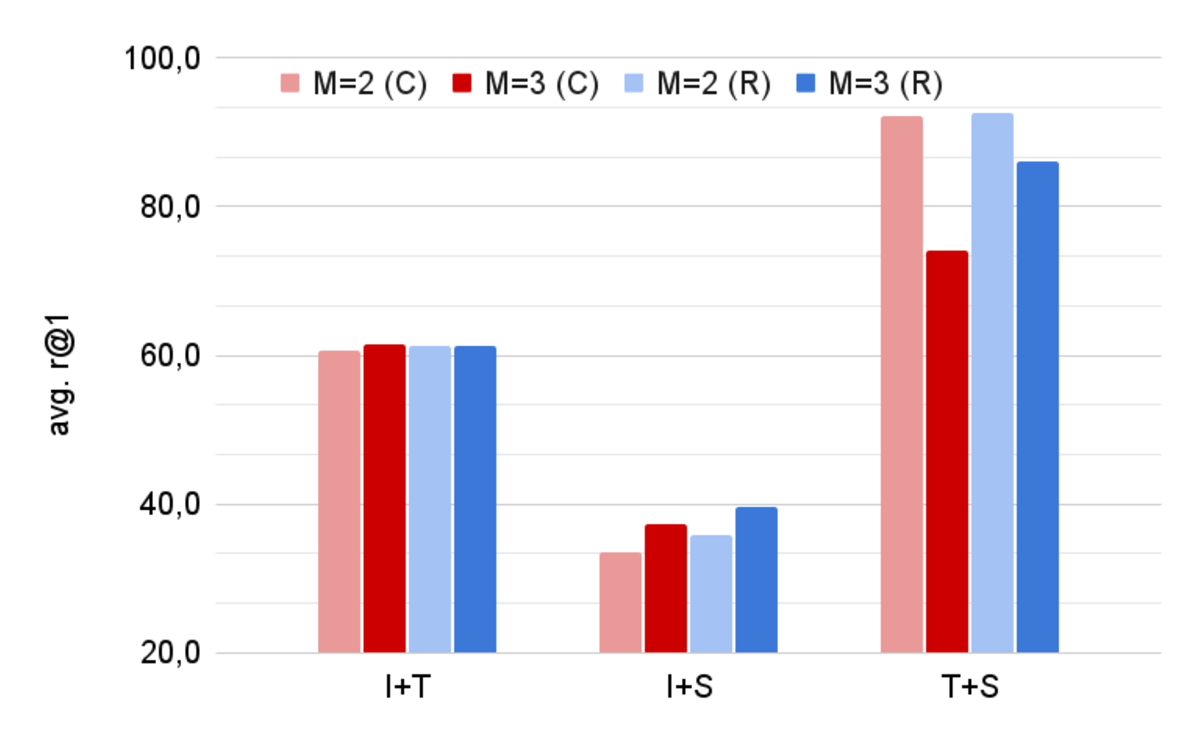}    
    \caption{Flickr8K (frozen backbones)\label{fig:n_mod_f8k}}
\end{subfigure}
\begin{subfigure}[b]{0.5\textwidth}
    \centering
    \includegraphics[width=0.9\textwidth]{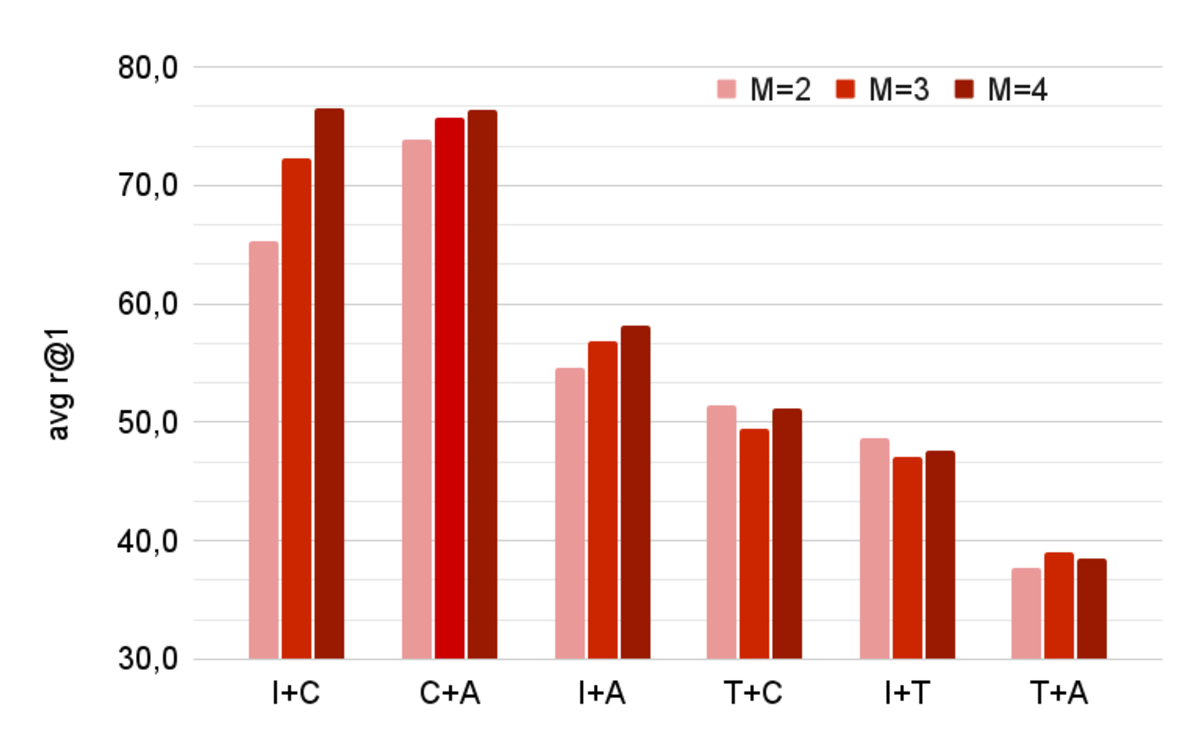}    
    \caption{CUB (fine-tuned backbones, contrastive loss)\label{fig:n_mod_cub}}
\end{subfigure}
\caption{Average cross-modal performance (avg. r@1) using $2,\dots,M$ modalities. Flickr8k: PCMC (C, red), PCMR (R, blue), frozen backbones. CUB: PCMC, fine-tuned backbones.}
\label{fig:n_mod}
\vspace{-1mm}
\end{figure}
In this section, we study the effect of using an increasing number of modalities for learning coordinated representations. \cref{fig:n_mod} illustrates average cross-modal performance for both Flickr8k (\cref{fig:n_mod_f8k}) and CUB (\cref{fig:n_mod_cub}) datasets. We show the average pairwise performance obtained by training a model using $2, \dots, M$ modalities. For example, on Flickr8k, performance on I+T for $M=2$ means we trained our models using only the image and language modalities, while for $M=3$ we trained coordinated representations using all three modalities (including speech) and then used only the image and language encoders for evaluation. 

For Flickr8k, we show pairwise cross-modal performance for both the PCMC (red) and PCMR (blue) settings, and for models trained using 2 and 3 modalities. We observe that for the combination of I+T (avg. of T$\rightarrow$I and T$\rightarrow$I) there is no noticeable gain in adding the speech modality. For the combination of I+S, learning with the extra text modality improves performance in both contrastive and non-contrastive cases. When considering the T+S combination, we see that for both PCMC and PCMR, adding the image modality to the mix seems detrimental compared to the bimodal setting. 
This could be explained by the differences in granularity observed between the text and speech compared to the image modality, \ie the fine-grained details that help disentangle similar caption and speech samples (those that describe the same image but differently) might be coarsened by forcing image samples to push them close to each other in embedding space, an effect that might show detrimental to the task.
Interestingly, the gap in performance observed between the bi- and tri-modal formulations is greater for the contrastive than for the regression-based one. 


For CUB, we observe a consistent increase for the I+C, C+A, and I+A combinations. For the rest, adding a third and fourth modality does not seem to give a consistent advantage over the bi-modal case. Interestingly, the observed performance drops occur for tasks that involve the text modality, which appear to be the most challenging ones as they exhibit the worst performance overall. We believe this shows the complexity of learning coordinated representations for modalities with different degrees of granularity, especially when combining coarse and fine-grained information. In both datasets, we observed little gains (if any) in adding extra modalities to the I+T combination, probably due to the a priori alignment of these two modalities (they are the image and text backbones of a pre-trained CLIP model). Nevertheless, we show that we are able to coordinate different types of input modalities in a unified and scalable manner.

\cref{tbl:cub_2to3} explores the effect of adding a third modality to a bi-modal setup under the PCMC loss and fine-tuned backbones. The table shows the relative gain in performance (avg. r@1) observed after adding a third modality to a bi-modal setup. The last column in the table shows the average improvement observed after adding each modality. As we observe, there is an overall positive effect of training models with additional modalities. However, the improvements depend on which modality is added in each case. For instance, we see that for problems involving T$\rightarrow$I and I$\rightarrow$T searches, adding class or attribute embeddings seems detrimental, perhaps due to the strong coupling between these modalities induced during pre-training.
{Also, as observed, adding the text modality brings consistent improvements on all pair-wise tasks.}.
\begin{table}
\small
\centering
\begin{tabular}{cccccccc}\toprule
mod. &I+C &C+A &I+A &T+C &I+T &T+A &avg. \\\midrule
+T &7.8 &1.9 &4.9 &- &- &- &4.9 \\
+I &- &3.1 &- &-7.0 &- &5.0 &0.4 \\
+A &13.6 &- &- &-0.4 &-5.3 &- &2.7 \\
+C &- &- &3.5 &- &-1.5 &2.3 &1.4 \\
\bottomrule
\end{tabular}
\caption{Average cross-modal improvement brought by training with an additional modality, when going from $M=2$ to $M=3$ on the CUB dataset, using PCMC and fine-tuned backbones.
}
\label{tbl:cub_2to3}
\vspace{-1mm}
\end{table}

\subsection{Zero-shot classification as cross-modal retrieval}

Given the flexibility of our approach 
, we look now at the problem of zero-shot classification. Although the cross-modal retrieval experiments in the previous section using the CUB dataset were carried out under a zero-shot setting, \ie using a disjoint set of training and test classes, we show that the advantages of our approach also translate to classification.

{We frame the classification task as a cross-modal retrieval problem by computing} embeddings for both the input and the output space (classes) and rank their similarity using the cosine metric. For the input space, we consider image (I), text descriptions (T), attributes (A), and their combinations. For the output space, we consider class embeddings (C) and text embeddings generated using the class name over a simple prompt (``A photo of a \{\}.") (P). \cref{{tbl:zsc_cub}} compares zero-shot performance (average per-class accuracy, T1) for different combinations of output and input embeddings, and compares them against different approaches from the literature. In our case, the combination 
{ operation consists of a simple average. For these experiments, we use PCMC with fine-tuned backbones.}

From the table, we see that using class embeddings generated by our model is way more effective than using simple textual prompts, as we observe considerable performance differences over all input embedding combinations except for the textual one, in which they perform on par. Class embeddings exhibit also better complementarity with the other input modalities. While the combination of image and attribute embeddings brings only a marginal improvement over using image embeddings alone in the case of prompt embeddings, it brings a +12\% boost in performance (64.4 for I vs. 72.3 for I+A) when encoding classes using learned projections. Interestingly, adding text descriptions to the mix does not bring any gain
{, which is consistent with the task (classification vs. regression) and the observations made in the previous section related to information granularity.}

Our approach compares also favorably to other methods from the literature. In the table, we report performance for different approaches that do not rely on feature generation, as this could also be used in conjunction with our approach to boost performance. SYNC \cite{changpinyo2016synnc} learns a mapping between the image and semantic space (class names or attribute embeddings) while preserving class-level relations. APN \cite{xu2020attribute} integrates local and global visual information using class-level attributes to regress local image representations. CD \cite{menon2022visual} ask GPT-3 for descriptive features for each class and use these descriptions as prompts to compute CLIP embeddings. DUET \cite{chen2023duet} encodes images and textual attributes using transformers and a cross-attention mechanism. Compared to the best-performing models (APN and DUET), our formulation led to a comparable classification performance when combining attribute and image embeddings for the input, and class embeddings for the output space.
\begin{table}
\centering
\small
\begin{tabular}{lcccccc}\toprule
\textbf{Model} &I &A &T &\multicolumn{2}{c}{\textbf{T1}} \\\midrule
SYNC \cite{changpinyo2016synnc} &\checkmark &\checkmark &- &\multicolumn{2}{c}{56.0} \\
APN \cite{xu2020attribute} &\checkmark &\checkmark &- &\multicolumn{2}{c}{72.0} \\
CD \cite{menon2022visual} &\checkmark &- &\checkmark &\multicolumn{2}{c}{65.3} \\
JE-ZSL \cite{nawaz2022semantically} &\checkmark &\checkmark &\checkmark &\multicolumn{2}{c}{54.1} \\
DUET \cite{chen2023duet} &\checkmark &\checkmark &- &\multicolumn{2}{c}{72.3} \\\midrule
& & & &C &P \\\midrule
CMPC &\checkmark &- &- &67.2 &34.3 \\
&- &\checkmark &- &57.7 &22.0 \\
&- &- &\checkmark &34.5 &34.5 \\
&\checkmark &\checkmark &- &72.9 &33.4 \\
&\checkmark &\checkmark &\checkmark &71.4 &30.4 \\
\bottomrule
\end{tabular}
\caption{Zero-shot classification performance on CUB for different combinations of input and output modalities, under the T1 metric \cite{xian2017gbu}. Columns 2-4 for the baseline models denote the modalities used to train each solution. For CMPC, these columns denote which of the modalities were aggregated to form the input representations. 
{We use PCMC and fine-tuned backbones on all four modalities.}
}
\label{tbl:zsc_cub}
\vspace{-1mm}
\end{table}

\subsection{Enriching the query for cross-modal retrieval}

In this section, we reconsider the cross-modal retrieval problem in the context of a more comprehensive multimodal setting and consider the effect of ``enriching" the query or database (DB) vectors with those from other modalities. In particular, we consider the image-to-text and text-to-image retrieval and evaluate different alternatives in which we complement the query (text or image) vectors with the information provided by other modalities (class and/or attribute embeddings). We focus on the image and text modalities since they are by far the most prevailing in the literature. Enriching the query vectors can be seen as a form of \emph{conditioning} (biasing the retrieval results towards the characteristics of the conditioning element) while doing it to the DB vectors, as a way to bias the representation towards some property of the data (\eg class structure) that is better aligned to the end task. \cref{tbl:enrichment} show cross-modal retrieval performance on the CUB and Flickr8k datasets for the (average) r@1 metric. The first two blocks of rows (\{\}$\rightarrow$X) denote query augmentation while the last two (X$\rightarrow$\{\}) database augmentation. The symbol \{\} must be understood as a placeholder to be filled by each modality combination shown in the adjacent columns. 

From the table, we see that in the case of CUB, enriching the query using either attribute or class embeddings provides a dramatic boost in retrieval performance. The improvement is larger for class embeddings since we are biasing the query towards the property that defines if a retrieved element is correct or not (we measure class-level r@1). Combining both attribute and class embeddings with an image/text query does not bring much compared to combining each of them separately. Similar observations can be made for the case of enriching the DB vectors. The difference is remarkable in the case of I$\rightarrow$T retrieval, as it improves over 24\%, 26\%, and 34\% after combining the DB vectors with attribute, class, and their combination. This is the only case in which the combination of attribute and class embeddings exhibit some complementarity. For Flickr8k, we observe that enriching the text modality is always detrimental, contrary to what happens when enriching the image modality with speech. Note the large improvement in the I+S$\rightarrow$T compared to I$\rightarrow$T. This could be explained by the tight alignment between text and speech. 
\begin{table}
\centering
\small
\begin{tabular}{lcccccc}\toprule
&\multicolumn{4}{c}{CUB} &\multicolumn{2}{c}{Flickr8k} \\\midrule
\multirow{2}{*}{\{\}$\rightarrow$T} &I &I+A &I+C &I+A+C &I &I+S \\
&56.9 &59.9 &67.5 &67.2 &66.8 &94.0 \\\midrule
\multirow{2}{*}{\{\}$\rightarrow$I} &T &T+A &T+C &T+A+C &T &T+S \\
&38.4 &60.6 &79.3 &80.3 &55.8 &51.6 \\\midrule
\multirow{2}{*}{T$\rightarrow$\{\}} &I &I+A &I+C &I+A+C &I &I+S \\
&38.4 &39.5 &40.2 &40.8 &55.8 &66.4 \\\midrule
\multirow{2}{*}{I$\rightarrow$\{\}} &T &T+A &T+C &T+A+C &T &T+S \\
&56.9 &70.7 &72.1 &76.3 &66.8 &62.5 \\
\bottomrule
\end{tabular}
\caption{Cross-retrieval performance (r@1) on CUB and Flickr8k by fusing the query (\{\}$\rightarrow$X) or DB vectors (X$\rightarrow$\{\}).}
\label{tbl:enrichment}
\vspace{-1mm}
\end{table}
\begin{figure}
\centering
\includegraphics[width=0.9\linewidth]{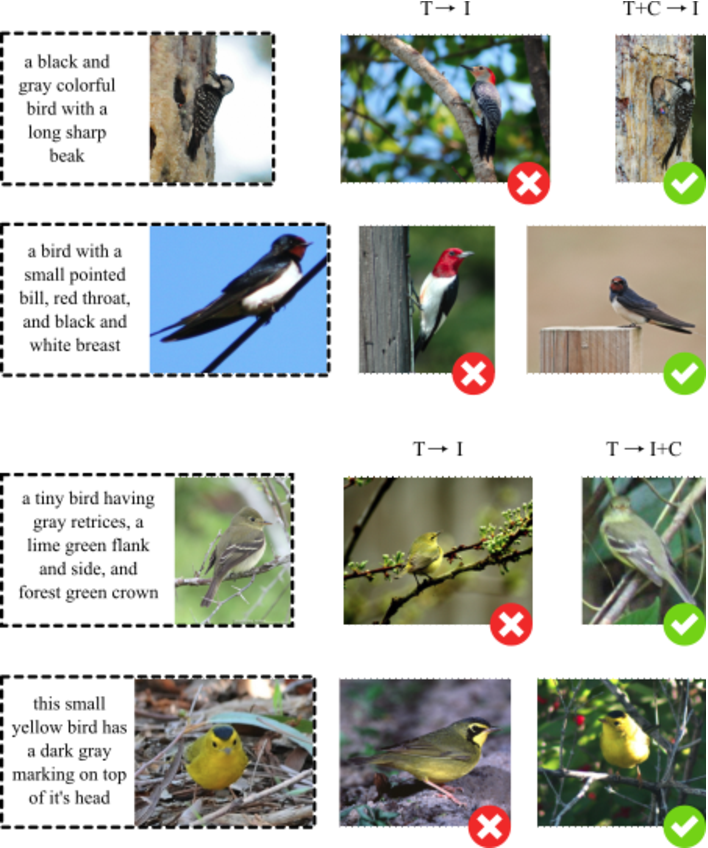}    
\caption{Qualitative cross-modal retrieval examples for enriched query (first two rows) and database vectors (last two rows). See text for details. Best viewed in color and with magnification.}
\label{fig:quali}
\vspace{-1mm}
\end{figure}

We provide some qualitative examples in \cref{fig:quali} for the case of text-to-image retrieval. The first two rows illustrate the effect of adding a class embedding to the text query (its embedding) and the last two the case of adding the corresponding class embeddings to the image embeddings stored in the database. The first element on each row shows the text caption being used to trigger the query. The image below (framed in dotted lines) corresponds to the matched image in the dataset and is shown only as a reference. This image is never used. The second column of each row shows the nearest cross-modal match under a cosine similarity metric. The third column shows how this mismatch can be corrected by enriching the query (first two rows) or database vectors (last two rows) using the corresponding class embeddings. 
From the examples, we see that both methods allow us to disambiguate rather challenging cases in which a simpler model fails. For the query enrichment, the overhead of adding an additional embedding is negligible compared to the cost of computing the cross-modal similarities. For the database case, this is a one-time operation that is paid while storing the representations to be retrieved.

%% file: sec/4_conclusions.tex
\section{Conclusions}
\label{sec:conclusions}

We proposed two different approaches to learning coordinated representations from a diverse set of modalities. Our approach is based on emphasizing the role of pairwise interactions during training. We show that the resulting models are able to compete and even surpass the performance of specialized bimodal models. Our experiments also show that by adding other modalities, we can extend the cross-modal retrieval to tackle problems like zero-shot classification while also helping disambiguate fine-grained retrieval tasks. We believe our work complements current trends in multimodal research and 
brings new ways to deal with a variety of problems.

%% file: sec/5_supplementary.tex
\clearpage
\onecolumn

\section{Cross-modal retrieval on CUB}
\cref{tbl:cub_full_rp} and ~\cref{tbl:cub_full_r1} show full cross-modal results for the experiments in \cref{sec:cross_modal_results} for the CUB dataset. Note that for CUB, there is an imbalance in the way cross-modal metrics are computed. For instance, for image-to-text retrieval (I$\rightarrow$T) each image has 10 different captions all of which are considered correct. However, for text-to-image (T$\rightarrow$I) there is only a single image that matches the (text) query. This imbalance is more noticeable in the r@1 score.
\begin{table*}[h]
\small
\centering
\begin{tabular}{lrrrrrrrrrrrrr}\toprule
\multirow{2}{*}{\textbf{Model}} &\multicolumn{12}{c}{\textbf{R-P}} \\\cmidrule{2-13}
&I$\rightarrow$T &T$\rightarrow$I &I$\rightarrow$C &C$\rightarrow$I &I$\rightarrow$A &A$\rightarrow$I &T$\rightarrow$C &C$\rightarrow$T &T$\rightarrow$A &A$\rightarrow$T &C$\rightarrow$A &A$\rightarrow$C \\\midrule
PCME \cite{chun2021probabilistic} &26.3 &26.8 &- &- &- &- &- &- &- &- &- &- \\
DAA \cite{li2022differentiable} &28.2 &28.5 &- &- &- &- &- &- &- &- &- &- \\
PCMDA \cite{wang2022paired} &30.0 &29.7 &- &- &- &- &- &- &- &- &- &- \\\midrule
PCMC & & & & & & & & & & & & \\
\hspace{10pt} frozen &25.1 &24.0 &62.6 &60.6 &37.1 &35.7 &26.2 &27.7 &19.1 &19.4 &51.9 &54.5 \\
\hspace{10pt} frozen image &29.7 &28.8 &63.9 &61.9 &37.5 &36.3 &34.7 &35.9 &24.0 &24.1 &54.2 &55.9 \\
\hspace{10pt} frozen text &25.3 &23.8 &66.3 &62.5 &37.9 &36.7 &25.9 &27.9 &19.0 &19.4 &53.5 &57.5 \\
\hspace{10pt} fine-tuned &30.2 &29.4 &67.4 &65.3 &39.1 &38.0 &34.7 &36.2 &23.8 &24.0 &55.0 &57.9 \\
PCMR & & & & & & & & & & & & \\
\hspace{10pt} frozen &25.1 &23.7 &50.0 &49.0 &36.9 &35.3 &23.4 &25.4 &19.8 &20.0 &49.3 &50.3 \\
\hspace{10pt} frozen image &30.1 &29.0 &50.9 &49.7 &37.1 &35.6 &30.9 &31.2 &24.7 &24.7 &50.0 &50.9 \\
\hspace{10pt} frozen text &25.1 &23.7 &48.9 &46.8 &36.6 &35.2 &23.6 &25.5 &19.6 &19.9 &50.2 &51.9 \\
\hspace{10pt} fine-tuned &30.4 &29.4 &48.0 &45.7 &36.6 &35.3 &30.7 &31.1 &24.6 &24.5 &49.1 &50.4 \\
\bottomrule
\end{tabular}
\caption{Cross-modal retrieval performance on the CUB dataset under the R-P score.}
\label{tbl:cub_full_rp}
\end{table*}
\begin{table*}[h]
\small
\centering
\begin{tabular}{lrrrrrrrrrrrrr}\toprule
\multirow{2}{*}{\textbf{Model}} &\multicolumn{12}{c}{\textbf{r@1}} \\\cmidrule{2-13}
&I$\rightarrow$T &T$\rightarrow$I &I$\rightarrow$C &C$\rightarrow$I &I$\rightarrow$A &A$\rightarrow$I &T$\rightarrow$C &C$\rightarrow$T &T$\rightarrow$A &A$\rightarrow$T &C$\rightarrow$A &A$\rightarrow$C \\\midrule
PCME \cite{chun2021probabilistic} &46.9 &35.2 &- &- &- &- &- &- &- &- &- &- \\
DAA \cite{li2022differentiable} &53.2 &37.7 &- &- &- &- &- &- &- &- &- &- \\
PCMDA \cite{wang2022paired} &52.7 &40.6 &- &- &- &- &- &- &- &- &- &- \\\midrule
PCMC & & & & & & & & & & & & \\
\hspace{10pt} frozen &48.0 &29.9 &62.3 &78.2 &57.3 &49.4 &26.0 &64.1 &25.9 &37.6 &90.2 &54.3 \\
\hspace{10pt} frozen image &54.4 &37.3 &63.7 &82.2 &59.3 &50.3 &34.5 &63.9 &34.4 &42.9 &92.2 &55.8 \\
\hspace{10pt} frozen text &51.5 &30.8 &66.0 &82.2 &61.8 &51.6 &25.8 &62.1 &25.5 &37.1 &90.2 &57.3 \\
\hspace{10pt} fine-tuned &56.9 &38.4 &67.1 &86.0 &63.0 &53.2 &34.6 &67.8 &34.3 &42.8 &93.9 &58.7 \\
PCMR & & & & & & & & & & & & \\
\hspace{10pt} frozen &47.5 &28.9 &49.9 &70.2 &57.2 &48.5 &23.3 &50.1 &25.8 &38.5 &92.2 &50.2 \\
\hspace{10pt} frozen image &53.5 &37.1 &50.8 &73.8 &57.4 &48.7 &30.8 &60.0 &34.2 &45.9 &88.1 &50.8 \\
\hspace{10pt} frozen text &51.7 &29.1 &48.8 &70.1 &59.8 &50.9 &23.5 &48.1 &25.5 &38.5 &88.1 &51.7 \\
\hspace{10pt} fine-tuned &57.1 &36.8 &47.8 &67.8 &60.7 &52.3 &30.6 &56.2 &34.7 &45.3 &86.1 &50.3 \\
\bottomrule
\end{tabular}
\caption{Cross-modal retrieval performance on the CUB dataset under the r@1 score.}
\label{tbl:cub_full_r1}
\end{table*}

\section{Density of samples in the embedding space}

\begin{figure}
\centering
\includegraphics[width=0.9\linewidth]{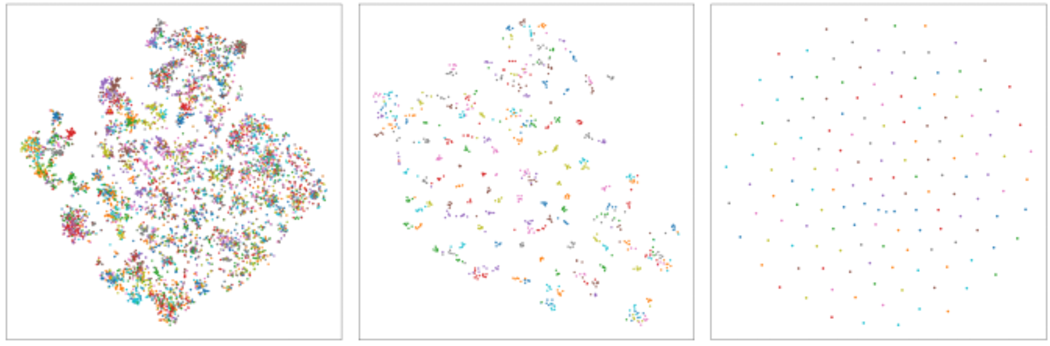}    
\caption{t-SNE projections of the text (left), image (middle), and class embeddings (right) learned on the CUB dataset using PCMC.}
\label{fig:tsne}
\end{figure}

\cref{fig:tsne} illustrates the difference between the different representations for a model trained on image, text, and class modalities. The figure shows 2D t-SNE projections for these modalities. As seen from the figure, the text modality has more variability than the image one, which is consistent with the nature of the CUB dataset, \ie free-form text captions describing close-caption images of 200 different bird species. The class modality can be seen in this case as well-separated class "prototypes".

